%% file: emnlp2023.tex
\pdfoutput=1
\PassOptionsToPackage{table}{xcolor}

\documentclass[11pt]{article}

\usepackage[]{EMNLP2023}

\usepackage{times}
\usepackage{latexsym}

\usepackage[T1]{fontenc}
\usepackage[utf8]{inputenc}

\usepackage{microtype}

\usepackage{inconsolata}

\input{new_commands_and_packages}

\newcommand{\ycom}[1]{{\color{black}{{#1}}}}
\newcommand{\rebut}[1]{{\color{black}{{#1}}}}

\definecolor{grayl}{gray}{0.9}
\usepackage{diagbox}

\title{Pointwise Mutual Information Based Metric and Decoding Strategy for Faithful Generation in Document Grounded Dialogs}

\author{Yatin Nandwani, Vineet Kumar, Dinesh Raghu, Sachindra Joshi  \and Luis A. Lastras \\
         IBM Research, AI \\
         \texttt{\{yatin.nandwani@, vineeku6@in, diraghu1@in, jsachind@in, lastrasl@us\}.ibm.com}
         }

\begin{document}
\maketitle

\input{pmi_decoding/sections/abstract}
\input{pmi_decoding/sections/introduction}

\input{pmi_decoding/sections/new_related_work}
\input{pmi_decoding/sections/methods}

\input{pmi_decoding/sections/experiments}
\input{pmi_decoding/sections/conculsion}
\input{pmi_decoding/sections/limitations}

\input{pmi_decoding/sections/ethics}
%\input{sections/acknowledgements}

\bibliography{anthology,custom}
\bibliographystyle{acl_natbib}
\appendix
\input{pmi_decoding/sections/appendix}

\end{document}

%% file: new_commands_and_packages.tex
\usepackage{bbm}
\usepackage{booktabs} 

\usepackage{lineno}
\usepackage{wrapfig}

\usepackage{comment}
\usepackage{amsmath}
\usepackage{comment}
\usepackage{soul}

\usepackage{multirow}
\usepackage{graphicx}
\usepackage{amssymb}
\usepackage{cleveref}
\usepackage{paralist}
\usepackage{todonotes}

\DeclareMathOperator*{\argmax}{arg\,max}

\newcommand{\bi}{\begin{itemize}}
\newcommand{\ei}{\end{itemize}}
\newcommand{\BE}{\begin{enumerate}}
\newcommand{\EE}{\end{enumerate}}

\newcommand{\eg}{\mbox{\it e.g.}}
\newcommand{\ie}{\mbox{\it i.e.}}

\newcommand{\wrt}{\mbox{\it w.r.t.}}

\newcommand{\viz}{\mbox{\it viz.}}

\newcommand{\qtwo}{Q$^2$}
\newcommand{\doc}{\textbf{d}}
\newcommand{\response}{\textbf{r}}
\newcommand{\responseword}{r}
\newcommand{\mybeta}{(1-\alpha)}
\newcommand{\responseseq}[1]{\langle\responseword_1\responseword_2 \ldots \responseword_{#1}\rangle}

\newcommand{\partialresponse}[1]{\response_{1:#1}}

\newcommand{\token}{v}

\newcommand{\vocab}{\mathcal{V}}

\newcommand{\history}{\textbf{h}}
\newcommand{\utterance}{u}
\newcommand{\turnm}{m}

\newcommand{\stept}{t}

\newcommand{\prob}{P}
\newcommand{\condprobfn}[2]{\prob(#1 | #2)}
\newcommand{\discreteprob}{p}

\newcommand{\probof}[2]{P_{#1}(#2)}

\newcommand{\faithfulnessmetric}{F}
\newcommand{\realspace}{\mathbbm{R}}
\newcommand{\pmimetric}{\text{\sc PMI--Faith}}
\newcommand{\upmimetric}{\text{\sc UPMI--Faith}}

\newcommand{\pmimetricbold}{{\sc \textbf{PMI--Faith}}}

\newcommand{\pmidecode}{\text{\sc PMI--Decode}}
\newcommand{\pmidecodebold}{{\sc \textbf{PMI--Decode}}}
\newcommand{\pmidecodeshort}{\text{PMI--D}}
\newcommand{\pmidecodenomask}{\text{PMI--D}_{\text{NM}}}
\newcommand{\pmidecodeeqwt}{\text{PMI--D}_{\text{EQ}}}

\newcommand{\pmi}{\text{PMI}}
\newcommand{\cpmi}{\text{CPMI}}

\newcommand{\mi}{\text{MI}}

\newcommand{\randvariablea}{A}
\newcommand{\randvaluea}{a}
\newcommand{\spacea}{\mathcal{A}}

\newcommand{\randvariableb}{B}
\newcommand{\randvalueb}{b}
\newcommand{\spaceb}{\mathcal{B}}

\newcommand{\randvariablec}{C}
\newcommand{\randvaluec}{c}
\newcommand{\spacec}{\mathcal{C}}

%% file: pmi_decoding/sections/abstract.tex
%Measuring the influence of knowledge in generating the response.
%Higher the influence, higher the fiathfulness.

\begin{abstract}

    A major concern in using deep learning based generative models for document-grounded dialogs is the potential generation of responses that are not \textit{faithful} to the underlying document. Existing automated metrics used for evaluating the faithfulness of response with respect to the grounding document measure the degree of similarity between the generated response and the document's content.
    However, these automated metrics are far from being well aligned with human judgments.
    Therefore, to improve the measurement of faithfulness, we propose a new metric that utilizes (Conditional) Point-wise Mutual Information (PMI) between the generated response and the source document, conditioned on the dialogue. PMI quantifies the extent to which the document influences the generated response -- with a higher PMI indicating a more faithful response. We build upon this idea to create a new decoding technique that incorporates PMI into the response generation process to predict more faithful responses. 
    Our experiments on the BEGIN benchmark demonstrate an improved correlation of our metric with human evaluation. We also show that our decoding technique is effective in generating more faithful responses when compared to standard decoding techniques on a set of publicly available document-grounded dialog datasets.

\end{abstract}

\begin{comment}
A major concern in using deep learning based generative models for document-grounded dialogs is the potential generation of responses that are not \textit{faithful} to the underlying document. Existing automated metrics used for evaluating the faithfulness of response with respect to the grounded document measure the degree of similarity between the generated response and the document's content. However, these automated metrics are far from being well aligned with human judgements. Therefore, to improve the measurement of faithfulness, we propose a new metric that utilizes (Conditional) Point-wise Mutual Information (PMI) between the generated response and the source document, conditioned on the dialogue. PMI quantifies the extent to which the document influences the generated response -- with a higher PMI indicating a more faithful response. We build upon this idea to create a new decoding technique that incorporates PMI into the response generation process to predict more faithful responses. Our experiments on the BEGIN benchmark demonstrate an improved correlation of our metric with human evaluation. We also show that our decoding technique is effective in generating more faithful responses when compared to standard decoding techniques on a set of publicly available document-grounded dialog datasets.

\end{comment}

%% file: pmi_decoding/sections/introduction.tex
\section{Introduction}

%Large Language Models (LLMs) have revolutionized the field of Natural Language Processing (NLP), particularly in complicated tasks related to text generation such as summarization \cite{}, paraphrasing \cite{}, and dialogue systems \cite{}. Dialogue systems can be classified into three main categories: open domain, task-oriented, and content-grounded. In this paper, we focus on content-grounded dialogue systems, in which the content for the conversation is either provided or retrieved from external sources.

Document--grounded dialog agents converse with users based on information present in document provided to them. These agents are expected to be factually consistent or \textit{faithful} to the grounding document and refrain from generating content that cannot be verified using the document. As most existing document--grounded dialog agents \cite{prabhumoye&al21,wu&al21} are built by fine-tuning large language models, ensuring faithful response generation is a major challenge.

To measure the ability of dialog agents to generate faithful responses, several automatic metrics have been proposed. These metrics take as input the agent generated response and the grounding document to quantify faithfulness. These are based on lexical overlap (e.g., BLEU, unigram-F1), semantic overlap (BERTScore) or even a trained classifier \cite{dziri&al22a}. Recently, 
%\citeauthor{honovich-etal-2021-q2} \citeyear{honovich-etal-2021-q2} 
\citet{honovich&al21} proposed Q$^2$, a metric that measures faithfullness using automatic question generation and question answering.

\begin{figure}[t]
    \centering
    \includegraphics[width=\columnwidth]{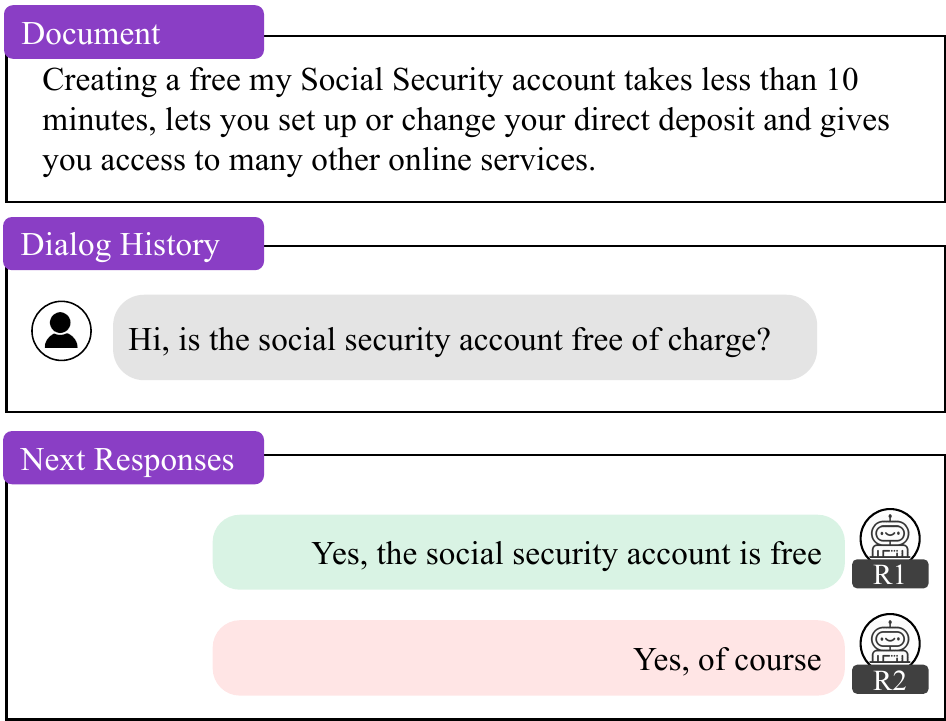}
    \caption{An example document grounded dialog with two types of responses: sentential response (R2) and non-sentential response (R1).}
    \label{fig:intro}
\end{figure} 

A major limitation of existing metrics is that they ignore the crucial dialog history when measuring faithfulness of responses. Even though, in many cases, the dialog history provides essential context that is necessary for a complete understanding of the response. To illustrate this point, let's consider two responses, textit{R1} and \textit{R2} , as depicted in Figure \ref{fig:intro}.
Response \textit{R1} is self-contained and can be comprehended without relying on the dialog history. On the other hand, response \textit{R2} is dependent on the dialog history and can only be fully understood when considering the preceding conversation. Unfortunately, current automated metrics do not take into account the dialog history, leading to their failure in evaluating responses that are not self-contained. Responses like $R2$ often lack domain-specific words, making similarity-based metrics like unigram-F1 and BERTScore ineffective. Additionally, generating question-answer pairs using such responses typically captures incomplete information, rendering metrics like Q2 as inadequate measures.

To overcome this problem, we propose a new metric that quantifies the faithfulness of a generated response with respect to both the document and the dialog history. Our metric is grounded in information theoretic concepts and captures the association of the response with the given document using Conditional Pointwise Mutual Information (CPMI).
We call our metric $\pmimetric$, which uses CPMI between the generated response and the document, conditioned on the dialogue history, for quantifying faithfulness. $\pmimetric$ captures the intuition that for a response to be grounded in the document, the probability of its generation given the document should be higher than the probability of its generation without the document.

A significant advantage of our metric $\pmimetric$ is that it can be  factorized the same way as the likelihood of a response can be factorized in auto regressive models. We take advantage of this property to propose a novel decoding objective, $\pmidecode$. The goal of $\pmidecode$ is to maximize not just the response's likelihood but a score that combines its likelihood and faithfulness. 

\noindent
To summarize, our contributions are threefold:
\begin{compactenum}
    \item We propose \pmimetric, a novel metric which quantifies faithfulness as a  conditional PMI between the response and the document given the dialog history.
    \item We propose a novel decoding objective, \pmidecode, which can aid in 
    %controlling the faithfulness of generated response. 
    generating faithful responses.
    \item Our experiments show that \pmimetric\ correlates with human judgments better than any existing metrics on the BEGIN benchmark \cite{dziri&al22}. We also show that using \pmidecode\ as the objective generates more faithful responses than standard likelihood objective on three standard document--grounded dialog datasets.
\end{compactenum}
We release our code~\footnote{\url{https://github.com/ynandwan/pmi-faith}} for further use by the research community.

%% file: pmi_decoding/sections/new_related_work.tex
\section{Related Work}
In this work, we focus primarily on faithfulness aspect of the generated responses with respect to the grounding document. It is crucial to distinguish between faithfulness and hallucination~\cite{Maynez2020OnFA} in evaluating responses. A response is considered faithful only when all the information it contains can be verified or inferred from the grounded document. On the other hand, a response is considered as a hallucination if it provides false or fabricated information. It is important to note that there can be responses that are not hallucinations but are still unfaithful. In such cases, the information provided may not be false, but it cannot be verified using the grounded document as a reference. 
%\st{It is important to point out that faithfulness is a subset of hallucination.} 
\rebut{It is important to point out that the set of faithful responses is a subset of responses that are not hallucinations.}
%When measuring faithfulness we are only concerned with the fact that generated response is attributable to the grounding document.
In this section, we discuss related work in faithfulness, \rebut{followed by a brief discussion on Mutual Information in conversational settings.}

Researchers have used various terms such as faithfulness \cite{cao&al18}, factual consistency \cite{cao&al20,santhanam&al21}, factual accuracy \cite{goodrich&al19}, fidelity \cite{chen&al20}, attribution \cite{rashkin&al21a} and hallucination (i.e., the lack of faithfulness) \cite{xiao&wang21} to define and quantify faithfulness of a model's generated text to a given knowledge.
%knowledge intensive NLP Tasks.

Most of the works focusing on evaluating faithfulness propose to train a classifier for the task \cite{goodrich&al19,kryscinski&al20,dziri&al22a}. Whereas our proposed metric doesn't require any training and is agnostic to the underlying data.
%In a recent work, \citet{dziri&al22a} fine-tune a RoBERTa based classifier to predict faithfulness in knowledge grounded dialogue systems. We compare it against our metric $\pmimetric$ that doesn't require any training and is agnostic to the underlying dataset.
%Most of the works focusing on evaluating faithfulness propose to train a classifier for the task. \todo{we can keep just the first line and cite all works there.} \citet{goodrich&al19} train a relation classifier, and  \citet{kryscinski&al20} fine--tune a BERT based classifier using weak supervision to predict factual accuracy in summarization task. In a recent work, \citet{dziri&al22a} fine-tune a RoBERTa based classifier, to predict faithfulness in knowledge grounded dialogue systems. They call their classifier \textit{faithcritic} and make it publicly available. In our experiments, we compare it against our metric $\pmimetric$ that doesn't require any training and is agnostic to the underlying dataset.
Recently, \citet{honovich&al21} proposed \qtwo\ for quantifying faithfulness. It uses a question generator to first generate question--answer (QA) pairs from the generated response. Then a QA system is used to find an answer, to the generated question, from the document. Finally, an NLI system is used to compare the two answers. %Unlike \citet{scialom&al21}, they use an NLI system to quantify faithfulness by comparing the two answers.
Though \qtwo\ uses the given document to check the faithfulness of a response, it ignores the dialog history. Thus, it may fail at handling responses that are non-self contained as depicted in \Cref{fig:intro}. Our metric $\pmimetric$ addresses this issue.

% \vspace{0.1ex}
% \noindent\textbf{Mitigating Hallucination:}
% Hallucination can be reduced by pre--processing data and removing instances from training data that are factually incorrect~\cite{shen&al21}. It can also be reduced by training control tokens and using them during response generation~\cite{filippova20,rashkin&al21b}. 
% We focus on decoding and do not require explicit changes to the training procedure or data.

%\cite{gabriel&al21} propose a generator--discriminator framework to generate faithful summaries by training the discriminator to re--rank different summaries based on faithfulness.

Many recent works \cite{dziri&al22, honovich&al22} have released different benchmarks that can be used to evaluate the performance of faithfulness metrics. 
While \citeauthor{honovich&al22} aim to standardize benchmark datasets across different generation tasks, \citeauthor{dziri&al22} focus on document--grounded dialogues, and thus we use their benchmark to compare our metric with various baselines. 

\rebut{
%Mutual information has been previously used in conversational settings. 
\citet{li&al16} and \citet{paranjape&manning21} use mutual information to prevent the conversational models from generating generic responses (such as “Sorry, I’m not sure about this topic”).
Contemporary to our work, \citet{ren&al23} propose to use a Conditional Pointwise Mutual Information (CPMI) based metric to evaluate relevance of a generated response with respect to a reference hypothesis for open-domain response generation. In contrast, our work is the first to use CPMI as a metric for evaluating the faithfulness of a response given a document and dialogue history in a document-grounded response generation.
}

%% file: pmi_decoding/sections/methods.tex
\input{pmi_decoding/sections/methods_background}

\section{Approach}
In this section, we describe our proposed metric for faithfulness -- \pmimetric. We then propose a decoding strategy \pmidecode\ based on our metric, with the objective of generating relevant and faithful responses.
%as compared to greedy or nucleus sampling \cite{}. 
\input{pmi_decoding/sections/methods_metric}

\input{pmi_decoding/sections/methods_decoding}

%% file: pmi_decoding/sections/methods_background.tex
\section{Background}
In this section, we first review the task of document-grounded dialog response generation, followed by the definition of the faithfulness metric.
%We then define metric to compute faithfulness of generated responses with respect to the document.

\vspace{1.0ex}
\noindent
\textbf{Document Grounded Response Generation:}
Let dialog history $\history = [\utterance_1, \cdots \utterance_{\turnm}]$ be a sequence of $m$ utterances in the dialog so far and $\doc$ be the document on which the next response is grounded. The task of  document-grounded dialog response generation
is to predict the next response, $\response = \responseseq{T}$, one token at a time, given the dialog history $\history$ and the document $\doc$. Here,  $\forall i, \responseword_i \in \vocab$, 
where $\vocab$ is the vocabulary of all possible tokens. 
The underlying model learns a probability distribution $\probof{}{\response | \doc,\history}$ over all possible responses $\response \in \vocab^+$,
\rebut{where $\vocab^+$ is the space of all the sequences having one or more tokens from vocabulary $\vocab$.}

Typically, this distribution is factorized over the tokens of $\response$ as:

\begin{equation}
    \probof{}{\response | \doc,\history} = \prod_{\stept=1}^T \probof{}{\responseword_{\stept} | \doc,\history, \partialresponse{\stept-1}}
    \label{eq:response-generation}
\end{equation}

%Typically, the selection of the next word $\responseword_\stept$ from $\probof{}{\responseword_{\stept} | \doc,\history, \responseword_{1:\stept-1}}$ is guided by the underlying decoding strategies such as greedy or nucleus sampling. 

%\vspace{1.2ex}
\noindent
\textbf{Faithfulness Metric:}
Most of the existing definitions (and metrics) for faithfulness focus mainly on document $\doc$ and response $\response$ but ignore the history $\history$ \cite{dziri&al22a, dziri&al22}. This may be for the sake of uniformity across different tasks such as summarization, grounded dialogue generation, and paraphrase generation. We qualify the definition of faithfulness specifically for the task of document--grounded dialogue generation.
Formally, a response $\response$ is considered `faithful' to a given document $\doc$ and the dialogue history $\history$ iff $\doc, \history \vDash \response$, where $\vDash$ represents logical entailment.

A faithfulness metric should quantify the faithfulness of the response $\response$ to the document $\doc$ and dialogue history $\history$. 
In general, such a metric should take $\response, \doc$ and $\history$ as its input and compute a score, $\faithfulnessmetric(\response, \doc, \history) \in \realspace$, such that a higher value of $\faithfulnessmetric(\response, \doc, \history)$ indicates a more faithful response.

%% file: pmi_decoding/sections/methods_metric.tex
\subsection{\pmimetricbold}
%\noindent \textbf{Point--wise Mutual Information:}
\begin{comment}
Let $\randvariablea$ and $\randvariableb$ represent two discrete random variables over discrete space $\spacea$ and $\spaceb$ respectively. 
Further, let $\randvaluea \in \spacea$ and $\randvalueb \in \spaceb$ represent their possible outcomes respectively.
With a little abuse of notation, we use function $\discreteprob()$ to represent an appropriate probability or conditional probability distribution function.
Then, Mutual Information between $\randvariablea$ and $\randvariableb$  captures the amount of information overlap between them, and is defined as expectation over Point--wise Mutual Information:
\begin{align*}
    \mi(\randvariablea; \randvariableb) =
    \sum\limits_{\randvaluea \in \spacea; \randvalueb \in \spaceb}
    \discreteprob(\randvaluea, \randvalueb)
    &\pmi(\randvaluea; \randvalueb)  \text{  where  } \\
    \pmi(\randvaluea; \randvalueb) 
    = 
    \log \frac{\discreteprob(\randvaluea, \randvalueb)}
              {\discreteprob(\randvaluea) \discreteprob(\randvalueb)} 
    =& \log \frac{\discreteprob(\randvalueb)\discreteprob(\randvaluea | \randvalueb)}
    {\discreteprob(\randvaluea) \discreteprob(\randvalueb)} \\
    =& \log \frac{\discreteprob(\randvaluea | \randvalueb)}
    {\discreteprob(\randvaluea)}
\end{align*}
Now, if $\randvariablec$ is another random variable, with $\randvaluec \in \spacec$ representing its possible outcome, one can define Conditional Point--wise Mutual Information (CPMI) between $\randvaluea$ and $\randvalueb$ given $\randvaluec$ as:
\end{comment}

\pmimetric\ is based on the information--theoretic concept of Pointwise Mutual Information.
\begin{comment}
If $\randvaluea, \randvalueb$ and $\randvaluec$ are three possible events, then one can define Conditional Point--wise Mutual Information (CPMI) between $\randvaluea$ and $\randvalueb$ given $\randvaluec$ as:
\begin{equation*}
    \cpmi(\randvaluea; \randvalueb | \randvaluec) = 
    \log \frac{\discreteprob(\randvaluea, \randvalueb | \randvaluec)}
              {\discreteprob(\randvaluea | \randvaluec) \discreteprob(\randvalueb |\randvaluec)}
              = 
    \log \frac{\discreteprob(\randvaluea | \randvalueb, \randvaluec)}
    {\discreteprob(\randvaluea | \randvaluec)}
\end{equation*}
\end{comment}
We use the notion of CPMI between generated response $\response$ and the document $\doc$ given the context $\history$ to capture the influence of the document in generating the response.  We define our metric, $\pmimetric$, for faithfulness of the response $\response$ to the document $\doc$ as:
\begin{align}
\label{eq:cpmidef}
    \nonumber &\pmimetric(\response, \doc, \history) =
    \cpmi(\response; \doc | \history) \\
    &=
    \log \frac
    {\condprobfn{\response, \doc}{\history}}
    {\condprobfn{\response}{\history} \condprobfn{\doc}{\history}} 
    %&=
    %\log \frac
    %{\condprobfn{\doc}{\history} \condprobfn{\response}{\doc,\history}}
    %{\condprobfn{\response}{\history} \condprobfn{\doc}{\history}}
    =
    \log \frac
    {\condprobfn{\response}{\doc,\history}}
    {\condprobfn{\response}{\history}} 
\end{align}

%Measuring $\pmimetric$ requires computing the conditional probability of generating the response given the document and history (numerator) and the conditional probability of generating the response given only the current context (denominator). 

%Intuitively, for a response to be grounded in the document, the probability of its generation given the document should be higher than the probability of its generation without the document. This is exactly what CPMI between the response and the document (given the dialogue history) captures. 
Mathematically, PMI is a measure of the strength of the association between two random events. A positive value of CPMI in \cref{eq:cpmidef}
%between the response $\response$ and the document $\doc$ given the dialogue history $\history$ 
implies that the probability of generating the response given the document and the dialogue history is higher than the probability of generating the response given only the dialogue history.
Hence, the response is likely to be grounded in the document.
On the other hand, if the response $\response$ is not faithful to the document $\doc$, the probability of its generation given the document and the dialogue history is likely to be similar to the probability of its generation without the document,
resulting in a lower value of $\pmimetric$.
We use pre--trained language models such as BLOOM \cite{bloom} or GPT2 \cite{radford&al19}, to compute these conditional probabilities ${\condprobfn{\response}{\doc,\history}}$ and ${\condprobfn{\response}{\history}}$.

%% file: pmi_decoding/sections/methods_decoding.tex
\subsection{\pmidecodebold}
\pmidecode\ is a decoding strategy whose objective is to generate responses that are both relevant and faithful. Typically, the goal of any decoding strategy is to select a response that has the maximum (log) likelihood:
\begin{equation}
    \label{eq:maxlikelihood}
    \response = \argmax\limits_{\tilde{\response} \in \vocab^+}
    \log \probof{}{\tilde{\response} | \doc, \history}
\end{equation}
The objective of \pmidecode\ is to select a response that is highly likely and faithful.
%to the given document and dialogue history. This can be achieved during inference by 
%selecting a response that 
This is achieved by maximizing a combination of likelihood and faithfulness quantified using an appropriate metric $\faithfulnessmetric$. With $\alpha \in [0,1]$, 
%$\mybeta = 1 - \alpha$, 
and a linear scoring function, we get:
\begin{equation}
\label{eq:maxscore}
%\begin{split}
    \response = \argmax\limits_{\tilde{\response} \in \vocab^+} \mybeta\log \probof{}{\tilde{\response} | \doc, \history} +
     \alpha\faithfulnessmetric(\tilde{\response},\doc,\history)
%\end{split}
\end{equation}
% \begin{align*}
%     \response =& \argmax\limits_{\tilde{\response} \in \vocab^+} \score(\tilde{\response},\doc,\history) \text{, where,  } \\
%     \score(\tilde{\response},\doc,\history) =& \mybeta\log \probof{}{\tilde{\response} | \doc, \history} + 
%     \alpha\faithfulnessmetric(\tilde{\response},\doc,\history)
% \end{align*}

With an auto--regressive model that 
%factorizes the probability distribution as in \cref{eq:response-generation}, and 
generates the response one token at a time, we use decoding strategies, such as greedy decoding,
\ycom{ beam search, nucleus sampling \cite{holtzman&al20}, or beam sampling} as a heuristic to find the maxima.
\ycom{
For ease of description,  we use the greedy decoding below, though our approach is agnostic to the choice of heuristic for maximising the objective function. It just modifies the standard log--likelihood objective with an additional term corresponding to faithfulness. Our choice of $\pmimetric$ as function $\faithfulnessmetric$ for quantification of faithfulness keeps the decoding heuristic tractable as shown below.
}

%in \cref{eq:maxlikelihood}. 
With \cref{eq:maxlikelihood} as the objective, greedy decoding would sample the next token $\responseword_{\stept}$ as follows:

\begin{align}
%\responseword_{\stept} = \argmax\limits_{\token \in \vocab}{}& \score(\responseseqwithvar{\stept-1}{\token},\doc,\history) \\
\responseword_{\stept} = {} & \argmax\limits_{\token \in \vocab}
 \log \probof{}{\partialresponse{\stept-1}, {\token} | \doc, \history} \nonumber \\
%{}& + \alpha\faithfulnessmetric(\responseseqwithvar{\stept-1}{\token}, \doc, \history)
= {} &\argmax\limits_{\token \in \vocab}
[\log \probof{}{\partialresponse{\stept-1} | \doc, \history} \nonumber \\ {} &+ \log\probof{}{\token| \doc, \history,\partialresponse{\stept-1}}] \label{eq:tokenscore}
%{}& +\alpha\faithfulnessmetric(\responseseqwithvar{\stept-1}{\token}, \doc, \history)
% \\
% \begin{split} 
%     = \argmax\limits_{\token \in \vocab}{}&
% \mybeta\log \probof{}{\token| \doc, \history, \partialresponse{\stept-1}}\\
% {}& +\alpha\faithfulnessmetric(\responseseqwithvar{\stept-1}{\token}, \doc, \history)
% \end{split}
\end{align}

\noindent In \cref{eq:tokenscore}, the likelihood term has been factorized 
%using \cref{eq:response-generation} 
and notice that its first term is independent of the next token candidate $\token$ and thus can be dropped while taking $\argmax$. Not all faithfulness metrics can be decomposed the same way as the likelihood term.  
%Now, when we use metrics like \qtwo\ or BERTScore as the faithfulness metric, finding the maxima by computing $\faithfulnessmetric$ for various possible partial responses at each step $\stept$ becomes computationally challenging.
One advantage of \pmimetric\ is that it can be decomposed the same way as likelihood as follows:
%\textcolor{red}{Too Long: On the other hand, when using $\pmimetric$ as the measure of faithfulness, one can decompose the $\pmimetric$ of partial response $\responseseqwithvar{\stept-1}{\token}$ into a sum of $\pmimetric$ of partial response $\responseseq{\stept-1}$, which is independent of $\token$, and token--level CPMI between the token $\token$ and document $\doc$ conditioned on not just the context $\history$ but also the partial response $\partialresponse{\stept-1}$: }
\begin{align}
    \begin{split} 
    \pmimetric&(\partialresponse{\stept-1},{\token},\doc,\history) \\
    = {} & \log \frac{\probof{}{\partialresponse{\stept-1}, {\token} | \doc,\history}}
             {\probof{}{\partialresponse{\stept-1}, {\token} | \history}}  \nonumber
    \end{split}\\
    \begin{split}
     = {} & \log \frac{\probof{} {\partialresponse{\stept-1} | \doc,\history}}
             {\probof{}{\partialresponse{\stept-1} | \history}} \\ \nonumber
     {} & + \log  \frac{\probof{}{\token | \doc,\history,\partialresponse{\stept-1}}}
             {\probof{}{\token | \history, \partialresponse{\stept-1}}}                 
    \end{split} \\
    \begin{split}
    \label{eq:pmibreakup}
     = {} & \pmimetric(\partialresponse{\stept-1},\doc,\history)\\
        {} & + \cpmi(\token;\doc | \history, \partialresponse{\stept-1})
    \end{split}
\end{align}
%Notice that the first term in \cref{eq:pmibreakup} is independent of $\token$ and thus it can also be dropped while taking $\argmax$. 

%We obtain \cref{eq:pmibreakupinlog} by using \cref{eq:response-generation} in \cref{eq:pmibreakupstart}.
By using $\pmimetric$ as $\faithfulnessmetric$ in \cref{eq:maxscore}, and dropping the two terms which are independent of $\token$ from \cref{eq:tokenscore} and \cref{eq:pmibreakup},  the objective of greedy decoding using the \pmidecode\ objective is expressed as:
\begin{equation}
\begin{split} \label{eq:pmitokenscore}
    \responseword_{\stept} =  \argmax\limits_{\token \in \vocab}{}&
    \mybeta\log\probof{}{\token| \doc, \history,\partialresponse{\stept-1}}\\
    {}& +\alpha\cpmi(\token;\doc | \history, \partialresponse{\stept-1})
\end{split} 
\end{equation}

To compute CPMI in \cref{eq:pmitokenscore},
the same language model can be used to get the conditional probabilities $\probof{}{\token | \doc, \history, \partialresponse{\stept-1}}$ and
$\probof{}{\token | \history, \partialresponse{\stept-1}}$ by separately passing $\doc$,$\history$, $\partialresponse{\stept-1}$ and $\history$, $\partialresponse{\stept-1}$, respectively, through the model.

We observed that using $\cpmi$ in the scoring function sometimes results in selecting tokens from the document which may interfere with the grammar. To mitigate this, instead of maximizing over the entire vocabulary $\vocab$ at each step $\stept$, we propose to maximize only over the `\textit{top p}' subset from the likelihood distribution, $\vocab_{p,\stept}$, defined as the minimum cardinality subset of tokens with the sum of their probabilities as $p$. We call this top $p$ masking:
\begin{equation}
\begin{split} \label{eq:pmitokenscorewithmask}
    \responseword_{\stept} =  \argmax\limits_{\token \in \vocab_{p,\stept}}{}&
    \mybeta\log\probof{}{\token| \doc, \history,\partialresponse{\stept-1}}\\
    {}& +\alpha\cpmi(\token;\doc | \history, \partialresponse{\stept-1})
\end{split} 
\end{equation}

The intuition here is that while CPMI has a positive influence on generating a more faithful response, it may negatively impact the grammatical structure. Therefore by restricting the vocabulary to $\vocab_{p,\stept}$, we use only highly probable tokens to form a response and thus are likely to generate responses that are faithful as well as grammatically correct.

%The intuition here is that when informative spans from the document are stitched together to form the response, they have to be grammatically correct and apt for the dialog context. CPMI has a positive influence on generating the informative spans, but a negative influence when generating the words for stitching them. By restricting the vocabulary to $\vocab_{p,\stept}$, we restrict the selection 
%to stitching words to the words with reasonable likelihood. So, when only stitching words have high likelihood, they automatically get selected, irrespective of their CPMI value.

%and reduce the influence of CPMI on them.

%% file: pmi_decoding/sections/experiments.tex
\section{Experiments}

Our experiments answer two research questions:
\begin{compactenum}
    \item \pmimetric: How does our novel metric perform when compared to exisitng metrics on a standard benchmark (\cref{sec:exp:metric})?
    \item \pmidecode: Does our proposed decoding technique generate responses that are more faithful compared to vanilla decoding techniques, while still maintaining relevance (\cref{sec:exp:decoding}).?
\end{compactenum}
%(1) to validate the ability of our metric, $\pmimetric$, to identify faithful responses , and (2) to demonstrate that our novel decoding strategy, $\pmidecode$, that optimizes a combination of likelihood and faithfulness indeed generates more faithful responses than greedy decoding that maximizes only the likelihood .
%For all our experiments, we use datasets, code and libraries that are publicly available under appropriate licenses for research use. We cite them wherever appropriate.

%For the latter, we fine-tune an LLM separately for three document-grounded dialog datasets and compare the responses generated by  $\pmidecode$ with responses from standard likelihood-based decoding.
%We first use various automated metrics for the comparison of faithfulness and relevance, including $\pmimetric$ and \qtwo, followed by a human study on randomly selected samples from two of the three datasets.

\input{pmi_decoding/sections/experiments_metric}

\input{pmi_decoding/sections/experiments_decoding}

%% file: pmi_decoding/sections/experiments_metric.tex
%We exploit an existing dataset of human annotations of faithfulness proposed by \citet{dziri&al22} and show that $\pmimetric$ performs better than existing metrics in identifying faithful responses. 

\subsection{\pmimetricbold: Experimental Setup}

\vspace{0.2ex}
\noindent \textbf{Dataset:}
We experiment using recently proposed BEGIN benchmark \cite{dziri&al22} for evaluating the ability of $\pmimetric$ to identify faithful responses.
This benchmark uses three document grounded datasets, \textit{viz}, CMU--DoG \cite{zhou&al18}, TopicalChat \cite{gopalakrishnan&al19}, and WoW \cite{dinan&al19}. It contains $11,059$ responses generated by three different models, GPT2 \cite{radford&al19}, DoHA \cite{prabhumoye&al21} and T5 \cite{raffel&al20}, on randomly selected samples from test splits of the 3 datasets. Each generated response is annotated by humans and classified into either `Fully--attributable', `Generic', or `Not fully--attributable'. Overall, 23.3\% of the total response have been classified as `Fully--attributable' by human annotators.

%In the test split, it has human annotation over $11,059$ responses generated by three different models (GPT2 \cite{}, DoHA \cite{} and T5 \cite{}) on randomly selected samples from test split of three different document grounded datasets, CMU--DoG \cite{}, TopicalChat \cite{}, and WoW \cite{}.
%Each response has been classified into three different classes: `Fully--attributable', `Generic', and `Not fully--attributable'. Overall, 23.3\% of the total response have been classified as `Fully--attributable' by human annotators.
\vspace{0.2ex}
\noindent
\textbf{Evaluation Metrics:} 
Our objective is to identify faithful responses. Accordingly, we consider `Fully--attributable' as the positive class and both `Generic' and `Not fully--attributable' as the negative class. 
\rebut{We use the same evaluation setup as recommended in \citet{honovich&al21}.}
For each metric, we first normalize the score using min-max normalization. The min and the max scores are identified from the dev set. We then identify an \rebut{optimum} threshold for each metric as the one that achieves the best F1 on the dev set. Finally, during test, we use the identified min, max and thresholds to classify a response as faithful. The thresholds, min and max for each metric are reported in Appendix \ref{app:minmax}. Once we have the predicted class, we then compute precision, recall, F1 score and accuracy achieved by each metric.
\rebut{As done in \citet{honovich&al21}, we also report calibration-free metrics that don't require any normalization. In addition to Spearman's and Pearson's correlation with human annotations, we also report AUROC for various faithfulness metrics.}

\vspace{0.2ex}
\noindent 
\textbf{Baselines:}
We compare our model against \qtwo, the state-of-the-art metric. We also compare against various lexical and semantic similarity-based, and trained classifiers as faithfulness metrics. Specifically, we use Unigram-F1 (U-F1), SacreBLEU\cite{matt18, papineni&al02}, and RougeL \cite{lin04} to capture lexical overlap between $\doc$ and generated response $\response$;  BERTScore\cite{zhang&al20} to capture $\response$'s semantic similarity with $\doc$. We use the code\footnote{https://github.com/orhonovich/q-squared} provided by \citet{honovich&al21} for all the above baselines.\footnote{For BERTScore, we change the underlying model to the current best \textit{microsoft/deberta-xlarge-mnli}.}
We also compare against \textit{faithcritic}\footnote{https://huggingface.co/McGill-NLP/roberta-large-faithcritic}\cite{dziri&al22a}, which is a pre--trained classifier to predict faithfulness of a response.

%for each metric.
%is identified as the one which maximizes F1 score over the devset.

%\input{pmi_decoding/tables/metric_performance_datasetwise}

\begin{figure*}[t]
    \centering
    \includegraphics[width=\textwidth]{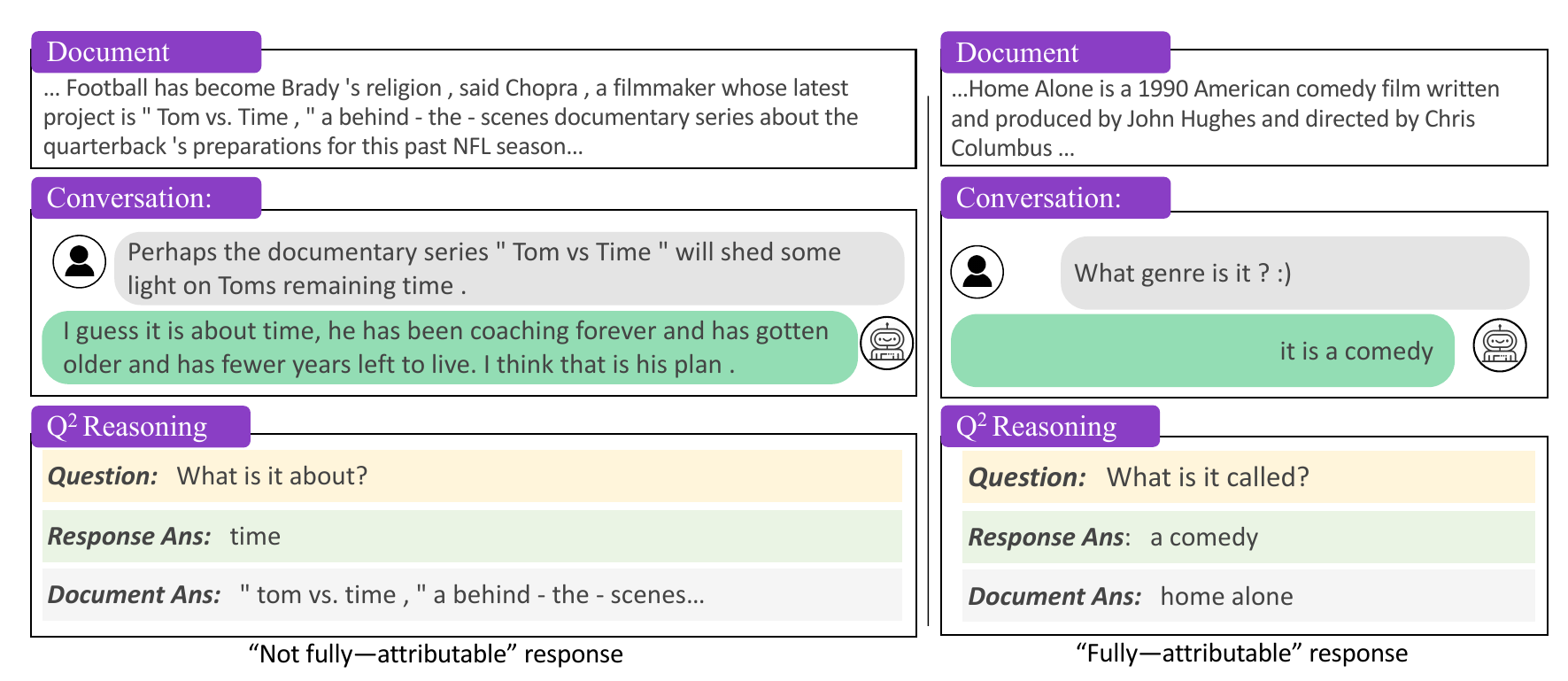}
    \caption{A `Fully attributable' (right) and a `Not fully attributable' (left) example, incorrectly classified by \qtwo\ and correctly classified by $\pmimetric$.}
    \label{fig:begin_ex}
\end{figure*} 

\vspace{0.2ex}
\noindent
\textbf{Training Details:}
To measure $\pmimetric$, we need to compute two conditional probabilities: $\probof{}{\response | \doc,\history}$, and 
$\probof{}{\response | \history}$.
To do so, we use pretrained LLMs available off the shelf from huggingface library \cite{wolf&al19}.
To quantify the impact of using one language model over the other, we compute the performance of $\pmimetric$ using eight LLMs of varying sizes: five BLOOM \cite{bloom} models with up to 7 billion parameters, and three GPT2 \cite{radford&al19} models up to GPT2-large (774 million).
We observe a robust and consistent performance with a variability of only $0.02$ points in the F1 score. Hence, for all further experiments, we use BLOOM-560m.
%, which is a 560 million parameter model, which unlike GPT2, has no restrictions on the number of input tokens.

\rebut{
\noindent
\textbf{Unconditional variant of $\pmimetricbold$:}
To quantify the impact of dialogue history $\history$ on \pmimetric, 
we also use a variant of it called unconditional PMI between a response and a document, \ie,  
$\upmimetric = \log \probof{}{\response | \doc} - \log \probof{}{\response}$,
to measure faithfulness.
%We call it Uncondonditional-PMI-Faith ($\upmimetric$)
%we used PMI, defined as  $log (P(r | d) / P(r) )$, as a metric for quantifying faithfulness and call it Uncondonditional-PMI-Faith (U-PMI-F). We observe that it indeed performs slightly inferior to PMIFaith (in terms of F1) and better than all other metrics. The tables below are a replica of Table 1 and Table 2 with an additional row for UncondPMIFaith:
}

\subsection{\pmimetricbold: Experimental Results}
\label{sec:exp:metric}

\input{pmi_decoding/tables/metric_performance_all}
\input{pmi_decoding/tables/calibration_free_metrics}
\Cref{tab:metric:all} reports the precision, recall, F1 score, and accuracy achieved by different metrics on the test split of BEGIN benchmark.
\rebut{We first observe that $\pmimetric$ performs better than $\upmimetric$, clearly demonstrating the advantage of using the dialogue history while measuring faithfulness.}

We \rebut{then} observe that \rebut{both $\upmimetric$ and} $\pmimetric$ perform better than all other faithfulness metrics by a considerable margin across all reported performance measures, with the absolute gains ranging from $21.8\%$ to $8.7\%$ in F1 score.
Even against the strong baseline of \qtwo, $\pmimetric$ achieves an absolute gain of $5.6\%$ and $8.7\%$ in accuracy and F1 score, respectively.
As expected, all the lexical overlap and semantic similarity based metrics achieve poor performance, with accuracy worse than even the majority--class classifier's accuracy of $76.7\%$.

Next, we notice that all metrics, except faithcritic, have higher recall than precision, indicating that they tend to be lenient while classifying a response as faithful, whereas faithcritic tends to be conservative and classifies most of the responses as not faithful.  Comparing the next two best metrics, we observe that \qtwo\ has better F1 and recall but worse accuracy and precision than faithcritic.

To identify dataset specific biases, \Cref{tab:metric:f1} reports F1 score separately for each of the three contributing datasets. 
We observe that $\pmimetric$ achieves the highest F1 on CMU--DoG and TopicalChat with more than  $12\%$ and $9.7\%$ absolute gain, respectively, over the other metrics.
FaithCritic achieves the best F1 score on WoW, whereas its F1 on TopicalChat and CMU--DoG is quite low.
This over-fitting on WoW is because faithcritic is a learned metric, and the training data for it has been adapted from WoW, and its low performance on the other two datasets demonstrates its lack of generalization. 

\rebut{
To understand the correlation of various faithfulness metrics with human judgement, we report three calibration-free metrics in \cref{tab:metric:calibrationfree}. 
 In all three metrics, we observe that PMI-Faith is better aligned to human judgements than the other measures of faithfulness.
}

\vspace{0.2ex}
\noindent \textbf{Subjective Analysis:}
 The state-of-the-art metric, \qtwo, identifies whether a response is faithful or not using two steps. In the first step, it generates a set of questions based on the response. In the second step, it uses a question answering system to generate two responses for each question: one based on the response and one based on the document. If both the answers match, then the response is considered faithful to the document. We now discuss the shortcoming of \qtwo and how \pmimetric\ overcomes it using two examples.

\Cref{fig:begin_ex} shows the two examples where $\pmimetric$ correctly identifies the faithfulness (or lack of it) whereas the strongest baseline \qtwo\ fails to do so. 
In the case of `Fully--attributable' response (right), the pronoun `\textit{it}' in the response is an anaphora, referring back to the antecedent `home alone' (movie name), which is difficult to infer without the dialogue context.
However, \qtwo\ doesn't take the dialogue history into account, and thus it considers the pronoun \textit{it} in the response as a cataphor, referring to its postcedent `comedy'. 
As a result, the QA system correctly answers the generated question `What is \textit{it} called?' with the postcedent `\textit{comedy}', when presented with the response, and correctly outputs its antecedent (\textit{home alone}) when presented with the document.
But the overall \qtwo\ system fails, as the two answers do not match.
On the other hand, by virtue of considering dialogue history during computation, $\pmimetric$ has information that the question is about the genre and not the movie name, and hence it can correctly classify the response as `Fully--attributable'.

The other example highlights two issues: (1) when the response is partially hallucinated, question generation system may generate question from just the faithful part of the response and may incorrectly declare the whole response as faithful. In this example, most of the response  contains an opinion, which is not faithful to the document, but the QG system focused on `\textit{I guess it is about time}'. 
(2): the other issue is that the NLI system fails to capture that the single word answer `\textit{time}' from the response is not entailed by the long answer from the document, resulting in the incorrect prediction by the overall system.
On the other hand, $\pmimetric$ considers the response as a whole, instead of separately focusing on parts of it. As a result, it is correctly able to identify the given response as not faithful. 
%\textcolor{red}{[this is too long; we need to shorten it]}

\input{pmi_decoding/tables/decoding_performance}

\input{pmi_decoding/tables/decoding_grid}

%% file: pmi_decoding/tables/metric_performance_all.tex
\begin{table}
\centering
\small
\begin{tabular}{@{}l | c c c c@{}}
\toprule
                     %& \multicolumn{4}{l}{\textbf{All Datasets}}                                 \\ \midrule
            \textbf{Metric}          & \textbf{Precision} & \textbf{Recall} & \textbf{F1}    & \textbf{Accuracy} \\ \midrule
{U-F1}        & 0.401              & 0.785           & 0.531          & 0.677             \\
{BLEU}        & 0.478              & 0.479           & 0.479          & 0.757             \\
{RougeL}      & 0.487              & 0.552           & 0.518          & 0.760             \\
{BERTScore}   & 0.459              & 0.673           & 0.546          & 0.739             \\
{FaithCritic} & \textbf{0.684}     & 0.492           & 0.573          & 0.829             \\
\qtwo                & 0.517              & 0.744           & 0.610          & 0.779             \\ \midrule
\upmimetric           & 0.592              & 0.704  & 0.643 & 0.818 \\
%| **U-PMI-F** | 59.2%         | 70.4%      | 64.3%     | 81.8%        |
\pmimetric           & 0.607              & \textbf{0.818}  & \textbf{0.697} & \textbf{0.834}    \\ \bottomrule
\end{tabular}
\caption{Performance of various faithfulness metrics on the BEGIN Benchmark.}
\label{tab:metric:all}
\end{table}

\begin{table}
\centering
\small
\begin{tabular}{@{}l | c c c@{}}
\toprule
                     & \textbf{CMU}   & \textbf{TC}    & \textbf{WoW}   \\ \midrule
%{U-F1}        & 0.462          & 0.429          & 0.587          \\
%{BLEU}        & 0.027          & -              & 0.620          \\
%{RougeL}      & 0.140          & 0.050          & 0.647          \\
{BERTScore}   & 0.292          & 0.432          & 0.612          \\
{FaithCritic} & 0.156          & 0.039          & \textbf{0.794} \\
\qtwo                & 0.543          & 0.487          & 0.705          \\ 
\midrule
%| **U-PMI-F** | 53.9%     | 54.2%     | 73.3%     |
\upmimetric           & {0.539} & {0.542} & 0.733          \\
\pmimetric           & \textbf{0.663} & \textbf{0.584} & 0.771          \\ \bottomrule
\end{tabular}
\caption{F1 score over each of the three contributing datasets in the BEGIN, \viz, CMU--DoG (CMU), TopicalChat (TC), and Wizards of Wikipedia (WoW). }
\label{tab:metric:f1}
\end{table}

%% file: pmi_decoding/tables/calibration_free_metrics.tex
\begin{table}[th]
\resizebox{1.0\columnwidth}{!}{
\begin{tabular}{@{}l |ccc@{}}
\toprule
\multicolumn{1}{l|}{\textbf{}} & \multicolumn{1}{l}{\textbf{Spearman Cor.}} & \multicolumn{1}{l}{\textbf{Pearson Cor.}} & \multicolumn{1}{l}{\textbf{AUROC}} \\ \midrule
{F1-U} & 0.449 & 0.447 & 0.804 \\
{BLEU} & 0.238 & 0.388 & 0.663 \\
{RougeL} & 0.412 & 0.465 & 0.781 \\
{BERTScore} & 0.433 & 0.479 & 0.796 \\
{FaithCritic} & 0.479 & 0.479 & 0.712 \\
{\qtwo} & 0.447 & 0.448 & 0.795 \\
\midrule
\upmimetric & 0.528 & 0.485	& 0.861 \\
{\pmimetric} & \textbf{0.597} & \textbf{0.650} & \textbf{0.907} \\ 
\bottomrule
\end{tabular}
}
\caption{Spearman and Pearson correlation with human annotations in the BEGIN Benchmark, and AUROC of various faithfulness metrics.}
\label{tab:metric:calibrationfree}
\end{table}

%% file: pmi_decoding/tables/decoding_performance.tex
\begin{table}
\footnotesize
\begin{tabular}{@{}l|l|cc|cc@{}}
\toprule
\multirow{2}{*}{\textbf{\begin{tabular}[c]{@{}l@{}}Decode\\ Method\end{tabular}}} &
  \multirow{2}{*}{\textbf{Obj.}} &
  \multicolumn{2}{c}{\textbf{Faithfulness}} &
  \multicolumn{2}{|c}{\textbf{Relevance}} \\ \cmidrule{3-6}
 &
   &
  \textbf{PMI-F} &
  \textbf{Q2} &
  \textbf{BLEU} &
  \textbf{RougeL} \\
  \midrule
 \rowcolor{grayl} \multicolumn{6}{c}{MultiDoc2Dial} \\ \midrule
\multirow{2}{*}{\begin{tabular}[c]{@{}l@{}}Beam \\ Search\end{tabular}} &
  Stand. &
  0.59 &
  0.63 &
  30.56 &
  0.488 \\  
 &
PMI-D & 0.64 &
  0.65 &
  28.95 &
  0.473 \\  \midrule
\multirow{2}{*}{\begin{tabular}[c]{@{}l@{}}Beam \\ Sampling\end{tabular}} &
  Stand. &
  {0.58} &
  {0.62} &
  {30.50} &
  {0.491} \\ 
 &
  PMI-D &
  {0.63} &
  {0.66} &
  {30.69} &
  {0.488} \\ \midrule
 \rowcolor{grayl} \multicolumn{6}{c}{TopicalChat} \\ \midrule
\multirow{2}{*}{\begin{tabular}[c]{@{}l@{}}Beam \\ Search\end{tabular}} &
  Stand. &
  0.49 &
  0.68 &
  6.63 &
  0.219 \\ 
 &
  PMI-D &
  0.57 &
  0.73 &
  5.65 &
  0.197 \\ \midrule
\multirow{2}{*}{\begin{tabular}[c]{@{}l@{}}Beam \\ Sampling\end{tabular}} &
  Stand. &
  {0.49} &
  {0.67} &
  {6.28} &
  {0.214} \\ 
 &
  PMI-D &
  {0.54} &
  {0.72} &
  {6.02} &
  {0.207} \\ \midrule
 \rowcolor{grayl} \multicolumn{6}{c}{FaithDial} \\ \midrule
\multirow{2}{*}{\begin{tabular}[c]{@{}l@{}}Beam \\ Search\end{tabular}} &
  Stand. &
  0.54 &
  0.83 &
  13.53 &
  0.404 \\ 
 &
  PMI-D &
  0.63 &
  0.87 &
  12.38 &
  0.389 \\ \midrule
\multirow{2}{*}{\begin{tabular}[c]{@{}l@{}}Beam \\ Sampling\end{tabular}} &
  Stand. &
  {0.52} &
  {0.82} &
  {13.30} &
  {0.398} \\ 
 &
  PMI-D &
  {0.61} &
  {0.87} &
  {12.38} &
  {0.388} \\ \bottomrule
\end{tabular}
\caption{\label{tab:results} Faithfulness and relevance metrics computed for various decoding techniques on three datasets.}
\end{table}

%% file: pmi_decoding/tables/decoding_grid.tex
\begin{table*}[ht]
\centering
\resizebox{0.8\textwidth}{!}{
\begin{tabular}{@{}|l|cccc|cccc|cccc|@{}}
\hline
 &
  \multicolumn{4}{c|}{\textbf{PMIF}} &
  \multicolumn{4}{c|}{\textbf{RougeL}} &
  \multicolumn{4}{c|}{\textbf{Grammatical errors (in \%)}} 
  \\ \hline
\diagbox{$p$}{$\alpha$} &
  {\textbf{0}} &
  {\textbf{0.25}} &
  {\textbf{0.5}} &
  \textbf{1} &
  {\textbf{0}} &
  {\textbf{0.25}} &
  {\textbf{0.5}} &
  \textbf{1} &
  {\textbf{0}} &
  {\textbf{0.25}} &
  {\textbf{0.5}} &
  \textbf{1} \\ \hline
\textbf{0.6} &
  \cellcolor{grayl}{0.52} &
  \cellcolor{grayl}{0.58} &
  {0.59} &
  0.59 &
  \cellcolor{grayl}{0.40} &
  \cellcolor{grayl}{0.40} &
  {0.39} &
  0.38 &
  6.9 & 10.2 & 11.8 & 12.2
  \\ 
\textbf{0.75} &
  \cellcolor{grayl}{0.52} &
  {0.60} &
  {0.61} &
  0.60 &
  \cellcolor{grayl}{0.40} &
  {0.39} &
  {0.38} &
  0.36 &
  %6.4%     |   12.0%  |  16.3%  | 18.6%
  6.4 & 12.0 & 16.3 & 18.6 \\ 
\textbf{0.9} &
  \cellcolor{grayl}{0.52} &
  {0.61} &
  {0.62} &
  0.57 &
  \cellcolor{grayl}{0.40} &
  {0.39} &
  {0.36} &
  0.30 &
   %7.5%     |   15.0%  |  24.2%  | 45.8%
  7.5 & 15.0 & 24.2 & 45.8 \\ 
\textbf{1} &
  \cellcolor{grayl}{0.52} &
  {0.61} &
  {0.56} &
  \cellcolor{pink}0.34 &
  \cellcolor{grayl}{0.40} &
  {0.38} &
  {0.30} &
  \cellcolor{pink}0.05 &
   %7.4%     |   16.5%  |  54.8%  |   -   | 
  7.4 & 16.5 & 54.8 & - \\ \hline
\end{tabular}%
}
\caption{ \label{tab:grid} Faithfulness, relevance, \rebut{and grammatical errors} of the responses generated by beam sampling using different configurations of $\alpha$ and $top-p$ masking on the FaithDial dataset}.
\end{table*}

%% file: pmi_decoding/sections/experiments_decoding.tex
\subsection{\pmidecodebold: Experimental Setup}

\noindent \textbf{Datasets:}
We perform our experiments on three document-grounded dialog datasets:
%We work with three datasets for PMI Decoding  Experiments
Multi-Doc2Dial~\cite{feng&al21}, TopicalChat
%\footnote{\url{https://github.com/alexa/Topical-Chat}}
~\cite{gopalakrishnan&al19} and FaithDial
%\footnote{\url{https://huggingface.co/datasets/McGill-NLP/FaithDial}}
~\cite{dziri&al22}. 
Each dialog in MultiDoc2Dial (MD2D) is between a user and an agent. Only the agent has access to the documents. So, we only model the agent responses for this dataset.
TopicalChat (TC) consists of dialogs between two parties, where each party may have a different set of documents on the same topics. We use the `rare'  version of the dataset and filtered utterances tagged as `personal knowledge'. FaithDial (FD), a faithful adaptation of WoW\cite{dinan&al19}, in which one participant can ask a wide range of questions and the other participant can only provide information from Wikipedia. Some statistics of the three datasets are in \Cref{tab:datasets}.
%for details.

\input{pmi_decoding/tables/decoding_data_stats}

\vspace{0.2ex}
\noindent
\textbf{Algorithms:} For each of the three datasets, we separately finetune a BART-Large \cite{lewis&al19}\footnote{\url{https://huggingface.co/facebook/bart-large}} model using the code\footnote{\url{https://github.com/McGill-NLP/FaithDial}} made available by \citet{dziri&al22a}. As baselines, we use two decoding techniques that use the standard likelihood as the objective function: (1) beam search and (2) beam sampling. Both the techniques use a beam size of 4. We compare these baselines with their variants that uses our \pmidecode\ ($\pmidecodeshort$) objective. We use the values of $\alpha$ and $\operatorname{top-p}$ masking that achieved the highest sum of RougeL and normalized \pmimetric\ on the dev set.

\begin{comment}
    
\vspace{0.2ex}
\noindent 
\textbf{Training Details:} For each of the three datasets, we separately finetune a BART-Large \cite{lewis&al19}\footnote{\url{https://huggingface.co/facebook/bart-large}} model using the code\footnote{\url{https://github.com/McGill-NLP/FaithDial}} made available by \citet{dziri&al22a}.
%in Table~\ref{tab:datasets}. 
Unless otherwise specified, we use the default parameter values.
%To fine-tune on Multi-Doc2dial, we employ a batch size of 16 for training, and train for 50 epochs with a patience of 20. 
%We fine-tune FaithDial using 4 gradient accumulation steps for 10 epochs. 
For both MD2D and TC, we use a batch size of 16 and train for 50 and 10 epochs, with patience of 20 and 5 respectively. For all three datasets, we use the model checkpoint which returns the best perplexity score on the dev set. Each model takes less than 14 hours of fine--tuning on an Nvidia A-100 GPU with 80GB memory.

We fine-tune a  LM separately for three document-grounded dialog datasets and compare the responses generated by  $\pmidecode$ with responses from standard likelihood-based decoding.
\ycom{In the experiments below, we use beam search with beam size 4 to optimize both the scoring functions.
}
\end{comment}

\subsection{\pmidecodebold: Experimental Results}
\label{sec:exp:decoding}

\input{pmi_decoding/tables/human-eval}
\input{pmi_decoding/tables/subjective_analysis_examples}

%\Cref{tab:results} shows various faithfulness and relevance metrics for different decoding strategies.
We compare the decoding strategies in \Cref{tab:results} using various automated metrics for faithfulness ($\pmimetric$, \qtwo) and relevance (BLEU, RougeL). The general trend is that the \pmidecode\ generates more faithful responses compared to the standard variant and the improvement in faithfulness comes a the cost of relevance. 

To gain a better understanding of the correlation between faithfulness and relevance, we conducted experiments involving different configurations of the control parameters of \pmidecodeshort\ ($\alpha$ and $\operatorname{top-p}$ masking). The results of these experiments are presented in the \cref{tab:grid}, showcasing the corresponding normalized PMIF and RougeL metrics for each configuration.

For a given fixed value of $\alpha$, as the $p$ value increases, the faithfulness of the responses also increases. However, this improvement in faithfulness comes at the cost of decreased relevance. On the other hand, for a fixed $p$ value, as $\alpha$ increases, the faithfulness initially increases and then gradually decreases. Simultaneously, the relevance decreases with an increase in $\alpha$. The highest level of faithfulness is observed when $\alpha=0.5$ and $p=0.6$. Meanwhile, the highest relevance is achieved when the CPMI is not used in the decoding objective ($\alpha=0$), or when both $\alpha$ and $p$ have low values.

We have gained two valuable insights from  \cref{tab:grid}. The first insight reveals that there are specific configurations, such as $p=0.6$ and $\alpha=0.25$, which achieve the same level of relevance (0.40) as the standard variant while generating more faithful responses. The second insight highlights a significant drop in relevance when solely focusing on PMI scores ($\alpha=1$ and $p=1$). Therefore, if the goal is to generate responses with nearly equivalent relevance, utilising smaller values for $\alpha$ along with a masking value of 0.6 is recommended.

\rebut{To demonstrate the impact of $\operatorname{top-p}$ masking on the grammar, we also report the percentage of grammatically incorrect responses for different values of $\alpha$ and $\operatorname{top-p}$ masking in \cref{tab:grid}. 
We use GECToR \cite{omelianchuk&al20}, a grammatical error correction method, to find if a generated response is grammatically correct or not.
We can easily see that for any value of $\alpha > 0$, the grammatical errors reduce significantly with a reduction in $top-p$. 
We do not report any value for $\alpha = 1$ and $top-p = 1$ as the responses with this configuration are not even in proper English. For example, one of the responses generated with $\alpha = 1$ and $top-p = 1$ is ``\textit{Cheyne Lauren sisters Vel Lauren wear Ralph indo Austrian Ralph linesauxricting Ren therapies Combat Rarity glamorous}".
}

%The second insight is that relevance gets hit by a large margin when you pay undue attention to just the PMI scores ($\alpha=1$) without masking. So, if you wish to generate responses that have almost the same relevance, then use small values for $alpha$ with masking set to 0.6.
\vspace{0.2ex}
\noindent \textbf{Human Evaluation:} We perform human evaluation experiments to compare (1) \textit{relevance}, (2) \textit{faithfulness}, and (3) \textit{grammar}. All three dimensions were categorically labeled as agree, neutral, or disagree. We sampled 100 random (document, dialog history, response) tuples, 50 each from MD2D and TC. We evaluate the responses generated by beam search using two objectives: standard and $\pmidecode$. Out of six in--house annotators used (3 per dataset), four were experts in dialog research and two were beginners. 
\rebut{Refer to \cref{appendix:human_eval} for more details.}

The results are summarized in Table \ref{tab:human-eval}. For each dimension, we report the percentage of responses that were rated \textit{agree}. As expected, \pmidecode\ generates more faithful responses compared to greedy. We observe a 15\% improvement in faithfulness compared to greedy decode on both datasets. Further, \pmidecode\ improves relevance on MD2D but slightly deteriorates on TC.
Manual analysis revealed that the improvement of relevance on MD2D is primarily due to inherent solution multiplicity~\cite{nandwani&al21} in most dialogues, where more than one correct response is possible, but the metrics capture just one.

As \pmidecode\ maximises not just the likelihood of responses, but a combination of likelihood and faithfulness, we expected the responses to contain grammatical errors compared to greedy decode. To counter this issue, we proposed to use a weighted combination of likelihood and faithfulness during decode, with a higher weight on likelihood. We also restricted the vocabulary during each decode step to just the \textit{top-p} subset. The human study shows that these mitigation techniques helped in reducing the grammatical mistakes made by \pmidecode. We see that the grammar is only slightly inferior to greedy on both the datasets.

We use Fleiss Kappa \cite{Fleiss1973TheEO} to measure the inter-annotator agreement, which is substantial for relevance (0.63) and faithfulness (0.63), and almost perfect (0.88) for grammar. 

\vspace{0.3ex}
\noindent \textbf{Subjective Analysis:} 
\Cref{tab:subjective-examples} presents an example from MD2D where standard likelihood based beam search decoding returns a generic response (`\textit{no info. found}') which is present in around $1800$ training samples. The same model returns the correct response when $\pmidecode$ objective is used instead of just likelihood, demonstrating the capability of $\pmidecode$ to shift the score in favour of the words present in the document. 

%% file: pmi_decoding/tables/decoding_data_stats.tex
\begin{table}
\centering
\resizebox{\columnwidth}{!}{%
\begin{tabular}{@{}lrrrrrr@{}}
\toprule
 & \multicolumn{3}{|c|}{\textbf{Num. samples}} & \multicolumn{3}{|c}{\textbf{Avg. words}} \\ \cmidrule{2-7}
\multicolumn{1}{l|}{} & \textbf{Train} & \textbf{Dev.} & \multicolumn{1}{r|}{\textbf{Test}} & \textbf{Doc.} & \textbf{Hist.} & \textbf{Resp.} \\ \midrule
\multicolumn{1}{l|}{\textbf{MD2D}} & 24,603 & 4,699 & \multicolumn{1}{r|}{4,567} & 166 & 93 & 18 \\
\multicolumn{1}{l|}{\textbf{TC}} & 131,555 & 8,183 & \multicolumn{1}{r|}{8,301} & 241 & 199 & 20 \\
\multicolumn{1}{l|}{\textbf{FD}} & 18,357 & 3,417 & \multicolumn{1}{r|}{3,539} & 23 & 69 & 18 \\ \bottomrule
\end{tabular}%
}
\caption{Various statistics of the three document grounded dialog datasets.
%MD2D: MultiDoc2Dial; TC: TopicalChat; FD: FaithDial.
}
\label{tab:datasets}
\end{table}

\begin{comment}
\begin{table}
\centering
\begin{tabular}{lccc}
\hline
\textbf{Dataset} & \textbf{Train} & \textbf{Dev} & \textbf{Test}\\
\hline
Multi-Doc2dial & 24,603 & 4,699 & 4,567 \\
TopicalChat & 131,555 & 8,183 & 8,301 \\
FaithDial &	18,357 & 3,417 &	3539\\
\hline
\end{tabular}
\caption{Number of train, development and test examples for all the datasets we used}
\label{tab:datasets}
\end{table}
\end{comment}

%% file: pmi_decoding/tables/human-eval.tex
\begin{table}
\centering
\small
\begin{tabular}{l|ccc|ccc}
\toprule
 & \multicolumn{3}{c}{\textbf{Multi-Doc2Dial}} & \multicolumn{3}{|c}{\textbf{Topical Chat}} \\
\cmidrule{2-7}
& Fai & Rel & Gra & Fai & Rel & Gra \\
\midrule
Standard & 0.52 & 0.72 & 0.96 & 0.69 & 0.70 & 0.96 \\
{PMI-D}  & 0.60 & 0.75 & 0.92  & 0.80 & 0.67 & 0.93 \\
\bottomrule
\end{tabular}
\caption{Human evaluation of responses generated using beam search with different decoding objectives. We evaluate faithfulness (Fai), relevance (Rel) and grammar (Gra). }
\label{tab:human-eval}
\end{table}

%% file: pmi_decoding/tables/subjective_analysis_examples.tex
\begin{table*}[ht]
\centering
\small
\begin{tabular}{l|p{13.7cm}}
\toprule
%\multicolumn{2}{c}{\textbf{Example 1}}\\
%\midrule
\textbf{Document} &
\textbf{What happens if I had a lapse of insurance coverage during active duty?} You are required to maintain continuous coverage or surrender your plates before deployed. If you were unable to do so , DMV has a procedure in place to exempt you from administrative actions upon your return. You must provide a notarized statement that the vehicle was not used during the time in question, a copy of your military ID, proof of current insurance or surrender of plates, and signed and dated copies of deployment and return papers, or DD-214 if separated from service...\\
\midrule
\textbf{Context} & \textbf{User}: lost my insurance while on active duty and have some questions\\
&\textbf{Agent}: You are required to maintain continuous coverage or surrender your plates before being deployed. Were you unable to keep your insurance while on active duty?\\
&\textbf{User}: yes, it just wasn't on my mind and I don't get notices and anything like that while deployed\\
%\midrule
%\textbf{Gold}& \textbf{Agent}: DMV has a procedure in place to exempt you from administrative actions upon your return. You must provide a notarized statement that the vehicle was not used during the time in question, a copy of your military ID, proof of current insurance or surrender of plates, and signed and dated copies of deployment and return papers, or DD-214. You can present these documents at any DMV office, or mail them to us.\\
\midrule
\textbf{Greedy} &\textbf{Agent}:  Unfortunately, no relevant information is found.\\
\midrule
\textbf{PMI}: & \textbf{Agent}: DMV has a procedure in place to exempt you from administrative actions upon your return. You must provide a notarized statement that the vehicle was not used during the time in question, a copy of your military ID , proof of current insurance or surrender of plates, and signed and dated copies of deployment and return papers , or DD-214 if separated from service.\\
\bottomrule
\end{tabular}
\caption{\label{tab:subjective-examples}
An example from the test set of Multi-Doc2Dial dataset where Greedy generates a `\textit{I don't know}' response and $\pmidecode$ generates a relevant and faithful response.}
\end{table*}

%% file: pmi_decoding/sections/conculsion.tex
\section{Conclusion}

In this paper, we present a novel metric, 
\pmimetric,
to measure faithfulness of responses generated by document grounded dialog systems. It uses conditional PMI between the response and the document given the dialog history to quantify faithfulness. We extend the idea of \pmimetric to propose a novel decoding objective,
\pmidecode\ which encourages responses to be faithful to the given document by maximizing both the likelihood and faithfulness of the decoded response. Our experiments on the BEGIN benchmark prove that our proposed metric better correlates with human judgments compared to existing metrics. On three document-grounded dialog datasets, our novel decoding objective generates more faithful responses than the standard likelihood objective, as measured using automated metrics and a human study.

%% file: pmi_decoding/sections/limitations.tex
\section*{Limitations}
Though our decoding objective generates more faithful responses, we observed its inability to respond to generic chit--chat or pleasantries, like `Hello!' or `Good--bye'.
It is possible to combine it with other techniques, like training with CTRL tokens \cite{rashkin&al21b}, which can enable it to generate both generic as well as faithful responses depending upon the dialogue context. But identifying when to generate a particular kind of response may require more insights and we leave this overall thread for future work.
Next, to compute $\cpmi$, we need to pass $\doc,\history$, and $\history$ separately to the decoder. Though it can be done in parallel, but it may still reduce the throughput of the overall system by half.
Finally, as demonstrated by the human evaluation, $\pmidecode$ at times generates grammatically incorrect responses, even though the pre--trained language models are very good at generating fluent and coherent English.
While we presented two knobs: $\alpha$ and $top \ p$ masking to overcome this, we believe there could be other ways of handling this.

\begin{comment}
ACL 2023 requires all submissions to have a section titled ``Limitations'', for discussing the limitations of the paper as a complement to the discussion of strengths in the main text. This section should occur after the conclusion, but before the references. It will not count towards the page limit.
The discussion of limitations is mandatory. Papers without a limitation section will be desk-rejected without review.

While we are open to different types of limitations, just mentioning that a set of results have been shown for English only probably does not reflect what we expect. 
Mentioning that the method works mostly for languages with limited morphology, like English, is a much better alternative.
In addition, limitations such as low scalability to long text, the requirement of large GPU resources, or other things that inspire crucial further investigation are welcome.
\end{comment}

%% file: pmi_decoding/sections/ethics.tex
\section*{Ethics Statement}
Our work does not introduce any new ethical concerns per se, other than the ones already faced by large language models.
Our decoding objective works on top of any trained language model and generates the text which is more faithful to a given input document.
This can act as a double--edged sword: on one hand,
if the document itself contains profanity, it may enhance the model's likelihood of generating similar content. 
But on the other hand, providing a valid document may also reduce the inherent likelihood of the model to generate profane content.
Therefore, we recommend using it with responsibility and caution.

\begin{comment}
Scientific work published at ACL 2023 must comply with the ACL Ethics Policy.\footnote{\url{https://www.aclweb.org/portal/content/acl-code-ethics}} We encourage all authors to include an explicit ethics statement on the broader impact of the work, or other ethical considerations after the conclusion but before the references. The ethics statement will not count toward the page limit (8 pages for long, 4 pages for short papers).
\end{comment}

%% file: pmi_decoding/sections/appendix.tex
\begin{comment}
\section{Ablation Study}

%\input{pmi_decoding/tables/appendix_ablation}

Table ~\ref{tab:ablation} shows the effect of changing $\alpha$ and $top\ p$ on a faithfulness and a relevance metric. We observe that the configuration with $alpha =1.0$ and no masking($top\ p = 1.0$) performs the worst in terms of both relevance and faithfulness. This highlights the need to condition the response on the document. We obtain the best balance between relevance and faithfulness with $alpha =1.0$ and $top\ p= 0.6$
\end{comment}

\section{Human Evaluation}
\label{appendix:human_eval}

\begin{figure*}[t]
    \centering
    \includegraphics[width=\linewidth]{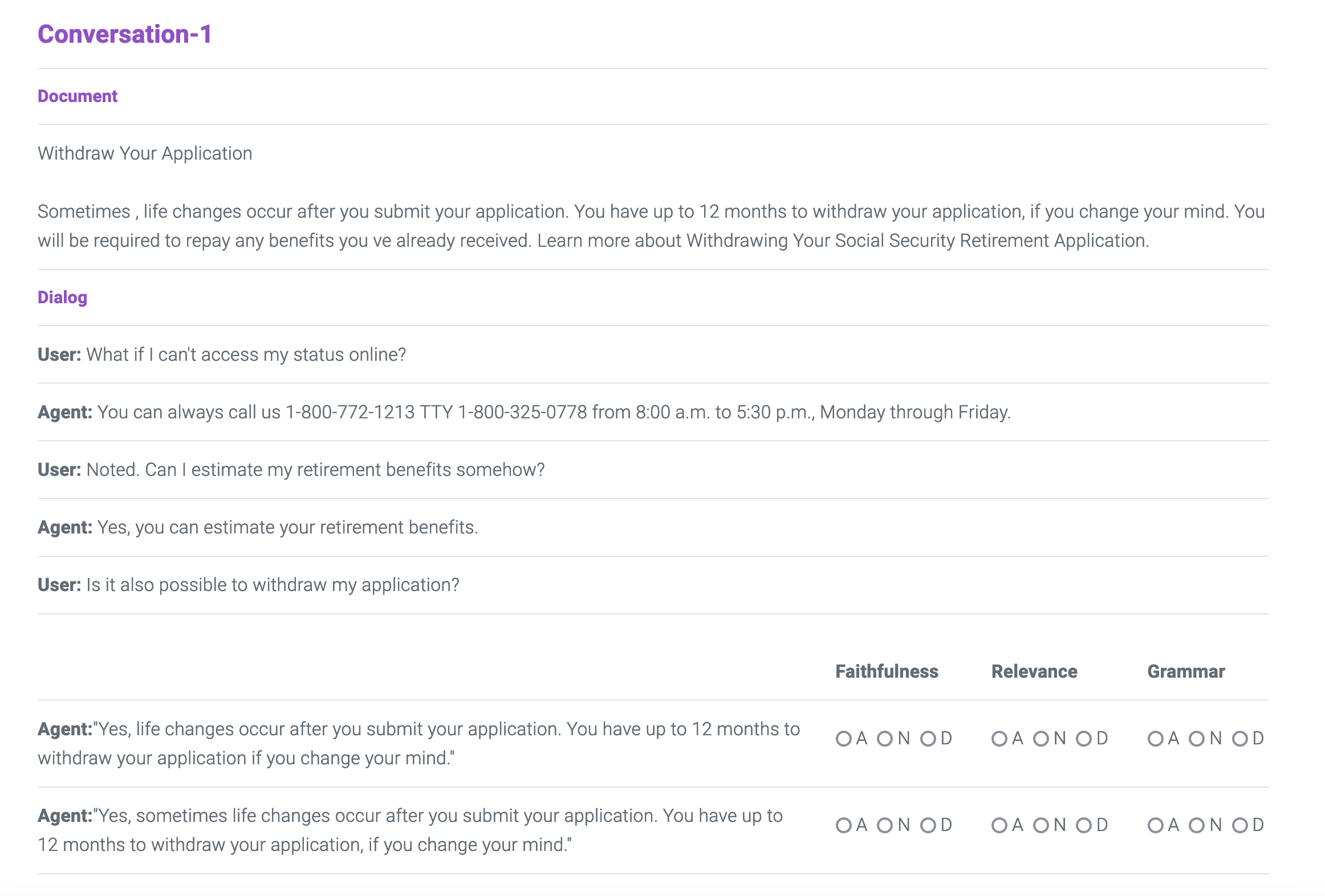}
    \caption{An example of the task provided to the human judges.}
    \label{fig:human-eval}
\end{figure*} 

The screenshot of a sample task is shown in Figure \ref{fig:human-eval}. 

\rebut{For our human evaluation study, we ensured the quality and expertise of our annotators by selecting individuals who are fluent in English and have a solid foundation in Machine Learning (ML), and Natural Language Processing (NLP).}
Out of six in--house annotators used, four were experts in dialog research and two were beginners.  
\rebut{Each annotator had completed at least one formal course in ML/NLP. The exact qualifications and experience of the six annotators are given below:

\begin{itemize}
    \item Annotator-1 is a postgraduate degree holder with more than 2 decades of experience in NLP research.
    \item Annotator-2 and annotator-3 are PhD degree holders with more than 5 years of experience in AI.
    \item Annotator-4 is an undergraduate degree holder with more than 5 years of experience in NLP research.
    \item Annotator-5 and annotator-6 are undergraduate degree holders with about 2 years of experience in AI research.
\end{itemize}
}
%As we used our in-house conversational AI experts for the experiments, we were only able to get it annotated for 2 datasets due to resource constraints and hence we randomly chose two datasets for this experiment.

All the annotators provided their consent over an appropriate official communication channel, \eg, official email or slack channels.

The following were the instructions provided to the human evaluators.

\noindent
\textbf{What is the task?} There are 50 incomplete dialogs along with a document over which the dialog is grounded on. For each (document, incomplete dialog) pair we provide the next response predicted by 2 different dialog systems (shuffled in random order). You are requested to judge the response generated by these 2 systems along three dimensions: faithfulness, relevance and grammar. Each dimension has to be evaluated using the following scale: Agree (A), Neutral (N), and Disagree (D).

\noindent
\textbf{How to judge relevance?}
Relevance measures how apt is the response given the dialog context and the knowledge.
Please select \textit{agree} when the response is apt and does not convey any incorrect information.  Select \textit{neutral} when it is hard to decide whether it is right or wrong and \textit{disagree} otherwise.

\noindent
\textbf{How to judge faithfulness?}
The faithfulness of a response is only dependent on the grounding document and it is independent of the dialog.
A system response can be marked disagree for relevance and still be marked agree for faithfulness.
Please select \textit{agree} when the complete response can be inferred from the document.  Select \textit{neutral} when it is hard to decide whether it can be inferred from the document or not and \textit{disagree} when a major portion of the response cannot be inferred from the document. For the case where the response is something like "No information is present". The judgement should be \textit{agree} if there is no information about that in the document provided and "disagree" if there is information available in the document, but the system didn't pick it up. For cases where the user initiates a chit-chat (say the user says "hi, how are you"), the agent responds with chit-chat ("I am doing good"), please can mark faithfulness as \textit{neutral}.

\noindent
\textbf{How to judge grammar?}
The grammar score for a response is independent of the dialog or the document.
A system response can be marked as disagree for relevance and still be marked agree for grammar.
Please select \textit{agree} when the response looks like how an expert human writes.  Select \textit{neutral} when there is a major issue with how the response reads but it still understandable and \textit{disagree} when the response makes no sense.

\section{Faithfulness Metric Normalization}
\label{app:minmax}
The minimum and maximium values used for normalizing various metrics are shown in Table \ref{tab:minmax}. These are the minimum and maximum values achieved by each metric on the dev set. We also report the threshold used on the normalized metrics for the faithfulness classification task. These are the thresholds than achieved the hightest F1 on the dev set.

% \usepackage{graphicx}
\begin{table}[h]
\centering
\small
\begin{tabular}{l|ccc}
\toprule 
            & \textbf{Min}   & \textbf{Max}  & \textbf{Threshold} \\ \midrule
U-F1        & 0     & 1    & 0.070     \\
BLEU        & 0     & 100  & 0.039     \\
RougeL      & 0     & 1    & 0.202      \\
BERTScore   & -0.639 & 1    & 0.440     \\
FaithCritic & 0     & 1    & 1        \\
\qtwo       & 0     & 1    & 0.625    \\
\upmimetric & -8.092     & 4.286    & 0.759    \\
\pmimetric  & -2.069 & 4.334 & 0.534    \\ \bottomrule
\end{tabular}%
\caption{\label{tab:minmax}. The min and max values used for normalizing each metric and the threshold used for the classifying faithfulness of a response using the metric.}
\end{table}